\documentclass{article}
\usepackage{amsmath,graphicx,epsfig}
\usepackage{url}

\usepackage{array,multirow}
 
\usepackage[preprint]{spconf}
\copyrightnotice{\copyright\ IEEE 2018}
\toappear{To appear in {\it Proc.\ ICIP 2018,
October 07-10, 2018, Athens, Greece}}

\usepackage{xcolor}

\title{Fully Convolutional Siamese Networks for Change Detection}
\name{Rodrigo Caye Daudt\textsuperscript{1,2}, Bertrand Le Saux\textsuperscript{1}, Alexandre Boulch\textsuperscript{1}}
\address{\textsuperscript{1}DTIS, ONERA, Universit\'{e} Paris-Saclay, FR-91123 Palaiseau, France \\ \textsuperscript{2}LTCI, T\'{e}l\'{e}com ParisTech, FR-75013 Paris, France}
\begin{document}
%
\maketitle
\begin{abstract}
This paper presents three fully convolutional neural network architectures which perform change detection using a pair of coregistered images. Most notably, we propose two Siamese extensions of fully convolutional networks which use heuristics about the current problem to achieve the best results in our tests on two open change detection datasets, using both RGB and multispectral images. We show that our system is able to learn from scratch using annotated change detection images. Our architectures achieve better performance than previously proposed methods, while being at least 500 times faster than related systems. This work is a step towards efficient processing of data from large scale Earth observation systems such as Copernicus or Landsat.
\end{abstract}
\section{Introduction}
\label{sec:intro}

Change Detection (CD) is one of the main problems in the area of Earth observation image analysis. Its study has a long history, and it has evolved alongside the areas of image processing and computer vision \cite{singh1989review,hussain2013change}. Change detection systems aim to assign a binary label per pixel based on a pair or sequence of coregistered images of a given region taken at different times. A positive label indicates the area corresponding to that pixel has changed between the acquisitions. While the definition of "change" may vary between applications, CD is a well defined classification problem. Changes may refer, for example, to vegetation changes, urban expansion, polar ice melting, etc. Change detection is a powerful tool in the production of maps depicting the evolution of land use, urban coverage, deforestation, and other multi-temporal types of analysis.

Programs such as Copernicus and Landsat make available large amounts of Earth observation imagery. These can be used in conjunction with advanced supervised machine learning algorithms that have been on the rise for the past decade, especially in the area of image analysis. It is therefore of interest to find efficient ways to make use of the available data. In the context of change detection there is a lack of large annotated datasets, which limits the complexity of the models that can be used. Nevertheless, there are available pixelwise annotated change detection datasets available that can be used to train supervised machine learning systems that detect changes in image pairs, such as the Onera Satellite Change Detection dataset\footnote{\url{http://dase.grss-ieee.org/}} presented in \cite{daudt2018urban} and the Air Change dataset~\cite{benedek2009change}. 



In this paper we present three Fully Convolutional Neural Network (FCNN) architectures that perform change detection on multi-temporal pairs of images Earth observation images. These architectures are trained end-to-end from scratch using only the available change detection datasets. The networks are tested in both the RGB and multispectral cases when possible. They are extensions of the ones presented in \cite{daudt2018urban}, which was the first CD method to be trained end-to-end, to a fully convolutional paradigm. This improves both the accuracy and the inference speed of the networks without increasing the training times. We present fully convolutional encoder-decoder networks that use the skip connections concept introduced in \cite{ronneberger2015u}. We also propose for the first time two fully convolutional Siamese architectures using skip connections.

This paper is organized as follows. Section~\ref{sec:related} explores other works that were used for inspiration or comparison during the development of this work. Section~\ref{sec:proposed} describes in detail the three proposed FCNN architectures for change detection. Section~\ref{sec:experiments} contains quantitative and qualitative comparisons with previous change detection methods. Finally, Section~\ref{sec:conclusion} contains this work's concluding remarks.


\section{Related work}
\label{sec:related}

\begin{figure*}
\begin{minipage}{0.19\linewidth}
\centering
\includegraphics[width=0.9\linewidth]{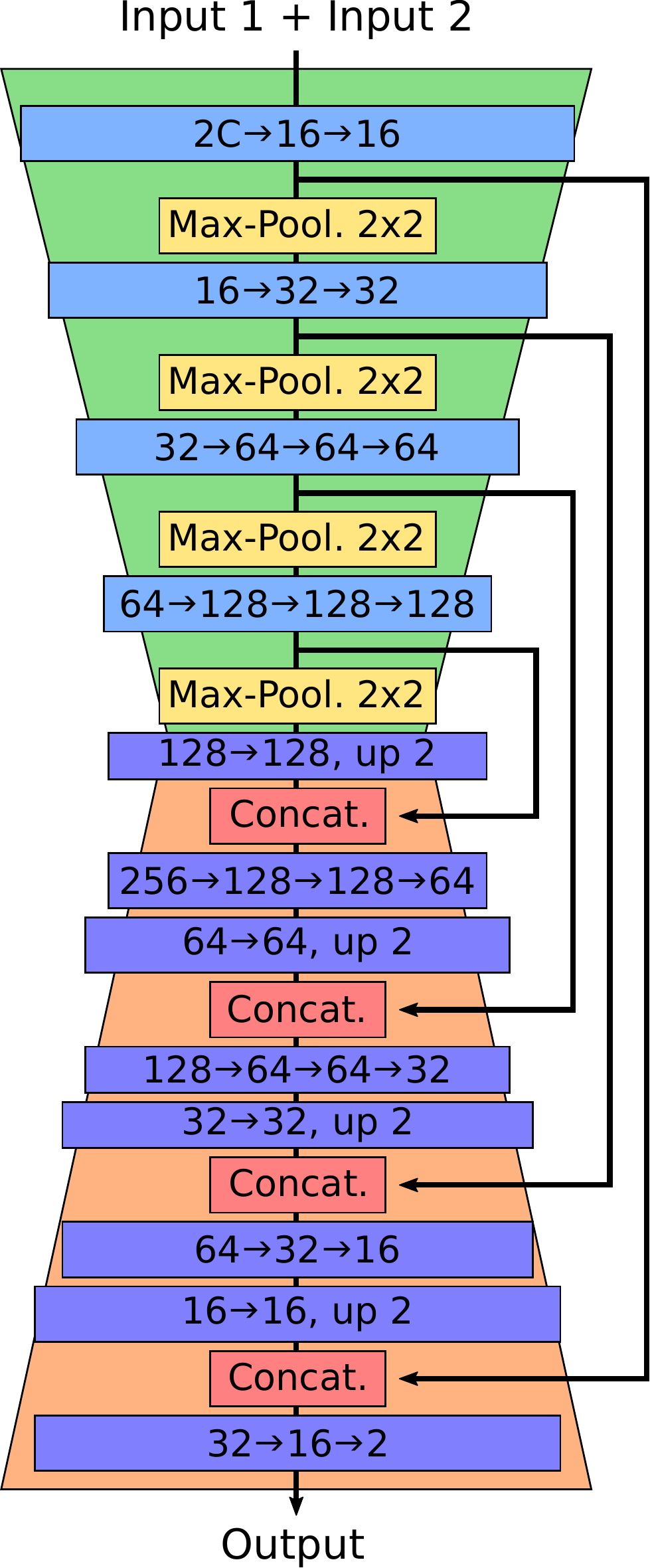}\\
(a) FC-EF.
\end{minipage}
\hfill
\begin{minipage}{0.39\linewidth}
\centering
\includegraphics[width=0.9\linewidth]{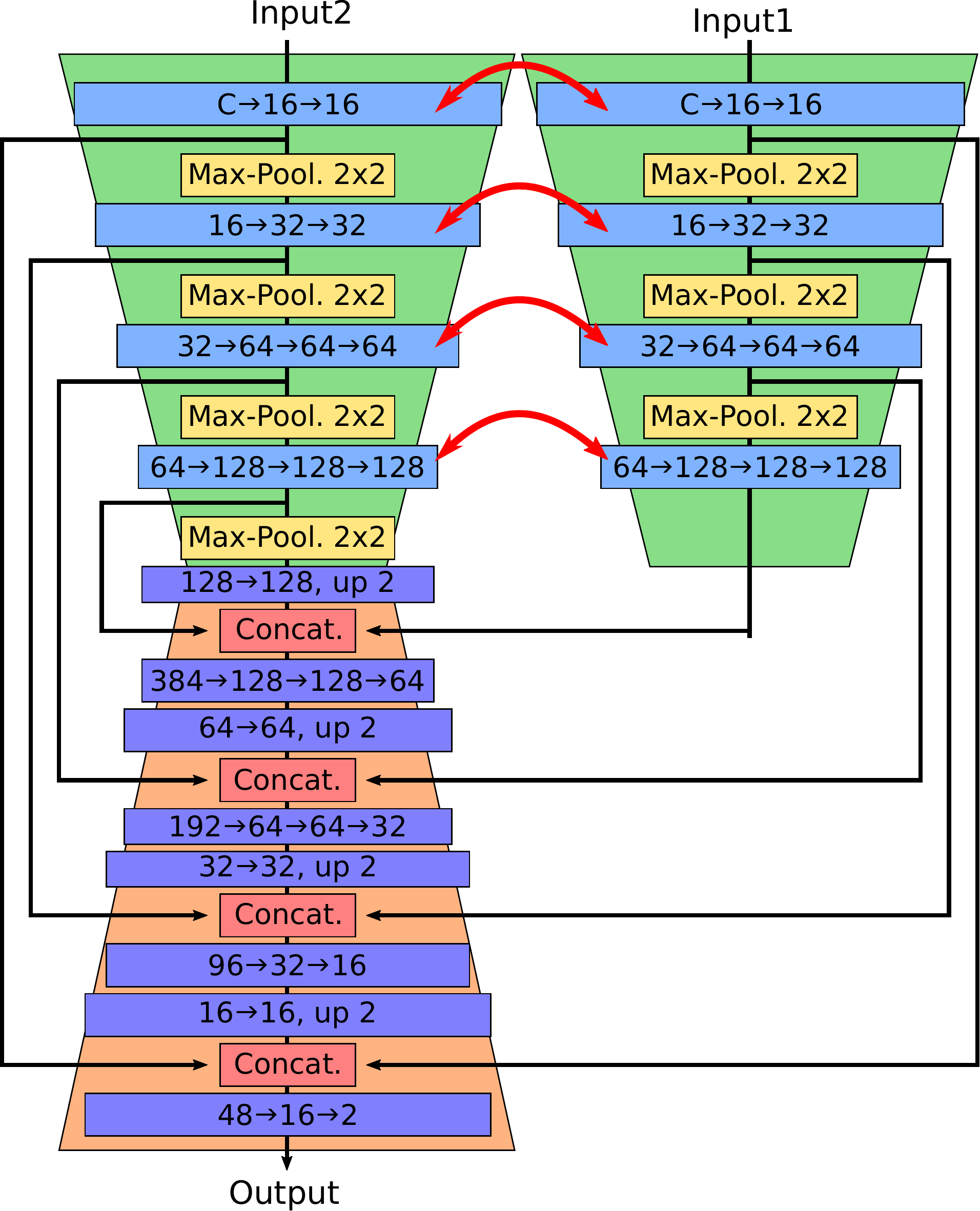}\\
(b)FC-Siam-conc.
\end{minipage}
\hfill
\begin{minipage}{0.40\linewidth}
\centering
\includegraphics[width=0.9\linewidth]{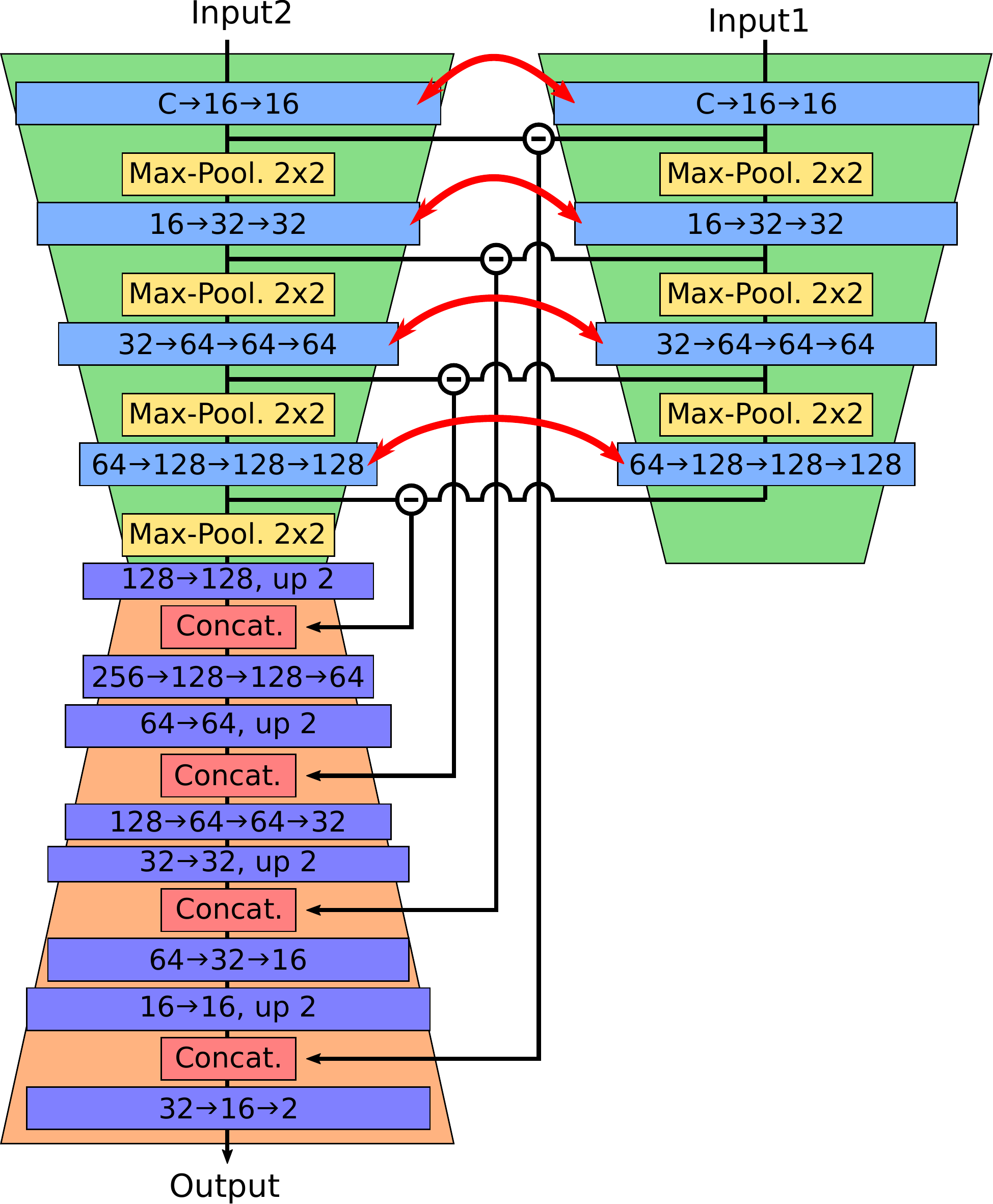}\\
(c) FC-Siam-diff.
\end{minipage}
\caption{Schematics of the three proposed architectures for change detection. Block color legend: blue is convolution, yellow is max pooling, red is concatenation, purple is transpose convolution. Red arrows illustrate shared weights.}
\label{fig:nets}
\end{figure*}

The history of change detection began not long after aerial images became possible~\cite{singh1989review,hussain2013change}. The proposed techniques have followed the tendencies of computer vision and image analysis: at first, pixels were analyzed directly using manually crafted techniques; later on, descriptors began to be used in conjunction with simple machine learning techniques~\cite{le2013urban}; recently, more elaborate machine learning techniques (deep learning) are dominating most problems in the image analysis field, and this evolution is slowly reaching the problem of change detection~\cite{stent2015detecting,liu2016deep,gong2016change,el2017zoom,zhan2017change,daudt2018urban,mou:hal-01561333}.

Due to the limited amount of available data, most of these methods use various techniques based on transfer learning, taking as a starting point networks that have been trained on larger datasets for different problems. This is limiting in many ways, as it assumes similarities between these datasets and the relevant change detection data. For example, most of the networks have been trained on RGB images, and cannot be transfered for SAR or multispectral images, which is the case of the dataset presented in \cite{daudt2018urban}. These methods also avoid end-to-end training, which tends to have better results for successfully trained systems. For this reason, the focus of this paper is algorithms that are able to learn solely from the available change detection data, and can therefore be applied to any available datasets.

Using machine learning for comparing images is not a new idea. Convolutional Neural Networks (CNNs) are a family of algorithms especially suited for image analysis, and have been applied in different contexts for the comparison of image pairs~\cite{chopra2005learning,zagoruyko2015learning,stent2015detecting}. Recently, fully convolutional architectures (FCNNs) have been proposed for problems that involve dense prediction, i.e. pixel-level prediction~\cite{long2015fully,ronneberger2015u,bertinetto2016fully}. Despite achieving state-of-the-art results in other Earth observation problems~\cite{audebert2017beyond}, these techniques have not yet been applied to CD to the best of our knowledge. The usefulness of such ideas in the context of Earth observation and their superiority over patch based, superpixel based and other approaches has already been studied~\cite{audebert2017segment}. Siamese architectures have also been proposed in different contexts with the aim of comparing images~\cite{bromley1994signature,zagoruyko2015learning}. To our knowledge, the only time a fully convolutional Siamese network was proposed previously was by Bertinetto et al. in \cite{bertinetto2016fully} to tackle the problem of object tracking in videos. Despite achieving good results, the architecture that was proposed in that work is specific to that problem and was not used as an inspiration to this work.






\section{Proposed approach}
\label{sec:proposed}

The work presented in this paper aimed to propose FCNN architectures able to learn to perform change detection solely from change detection datasets without any sort of pre-training or transfer learning from other datasets. These architectures are able to be trained end-to-end, unlike the majority of recent works on change detection. These fully convolutional architectures are an evolution of the work presented in \cite{daudt2018urban}, where a patch based approach was used. Moving the patch-based architectures to a fully convolutional scheme improves accuracy and speed of inference without affecting significantly the training times. These fully convolutional networks are also able to process inputs of any sizes given enough memory is available.

Two CNN architectures were compared in \cite{daudt2018urban}: Early Fusion (EF) and Siamese (Siam). The EF architecture concatenated the two patches before passing them through the network, treating them as different color channels. The Siamese architecture processed both images separately at first by identical branches of the network with shared structure and parameters, merging the two branches only after the convolutional layers of the network. 

To extend these ideas we used the concept of skip connections that were used to build the U-Net, which aimed to perform semantic segmentation of images~\cite{ronneberger2015u}. In summary, skip connections are links between layers at the same subsampling scale before and after the encoding part of an encoder-decoder architecture. The motivation for this is to complement the more abstract and less localized information of the encoded information with the spatial details that are present in the earlier layers of the network to produce accurate class prediction with precise boundaries in the output image.

The first proposed architecture is directly based on the U-Net model, and was named Fully Convolutional Early Fusion (FC-EF). We adapted the U-Net model into the FC-EF taking into account the amount of available training data. The FC-EF (Fig.~\ref{fig:nets}(a)) contains therefore only four max pooling and four upsampling layers, instead of the five present in the U-Net model. The layers in FC-EF are also shallower than their U-Net equivalents. As in the patch based EF model, the input of this network is the concatenation the two images in the pair that is to be compared.


The two other proposed architectures are Siamese extensions of the FC-EF model. To do so, the encoding layers of the network are separated into two streams of equal structure with shared weights as in a traditional Siamese network. Each image is given to one of these equal streams. The difference between the two architectures is only in how the skip connections are done. The first and more intuitive way of doing that is concatenating the two skip connections during the decoding steps, each one coming from one encoding stream. This approach was named Fully Convolutional Siamese - Concatenation (FC-Siam-conc, Fig.~\ref{fig:nets}(b)). Since in CD we are trying to detect differences between the two images, this heuristic was used to combine the skip connections in a different way. Instead of concatenating both connections from the encoding streams, we instead concatenate the absolute value of their difference. This approach was named Fully Convolutional Siamese - Difference (FC-Siam-diff, Fig.~\ref{fig:nets}(c)).

\section{Experiments}
\label{sec:experiments}

\begin{figure*}[ht]

  \begin{minipage}[b]{0.16\linewidth}
    \centering
    \centerline{\epsfig{figure=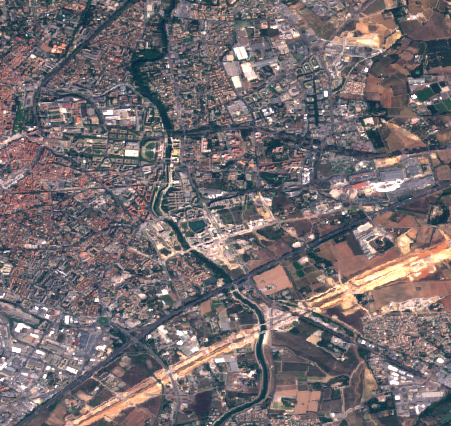,width=0.8\linewidth}}
    \centerline{(a) Image 1.}\medskip
  \end{minipage}
  \hfill
  \begin{minipage}[b]{0.16\linewidth}
    \centering
    \centerline{\epsfig{figure=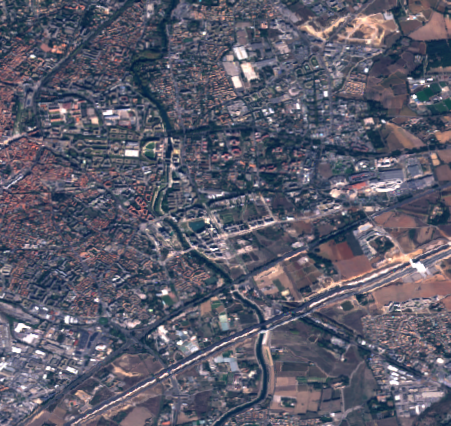,width=0.8\linewidth}}
    \centerline{(b) Image 2.}\medskip
  \end{minipage}
  \hfill
  \begin{minipage}[b]{0.16\linewidth}
    \centering
    \centerline{\epsfig{figure=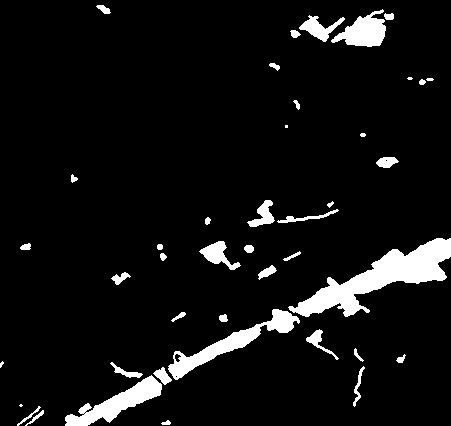,width=0.8\linewidth}}
    \centerline{(c) GT.}\medskip
  \end{minipage}
  \hfill
  \begin{minipage}[b]{0.16\linewidth}
    \centering
    \centerline{\epsfig{figure=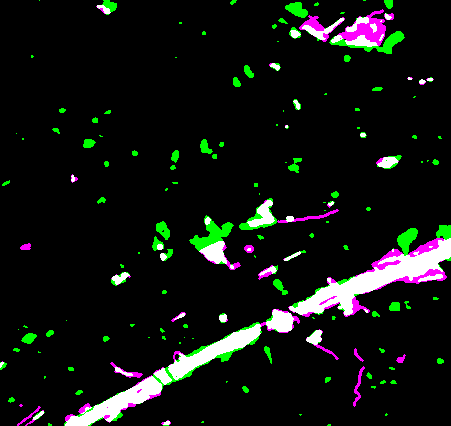,width=0.8\linewidth}}
    \centerline{(d) FC-EF.}\medskip
  \end{minipage}
  \hfill
  \begin{minipage}[b]{0.16\linewidth}
    \centering
    \centerline{\epsfig{figure=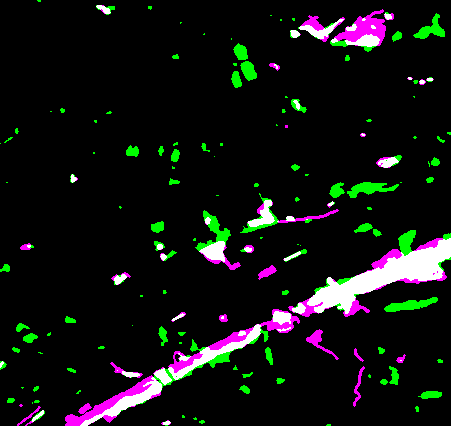,width=0.8\linewidth}}
    \centerline{(e) FC-Siam-conc.}\medskip
  \end{minipage}
  \hfill
  \begin{minipage}[b]{0.16\linewidth}
    \centering
    \centerline{\epsfig{figure=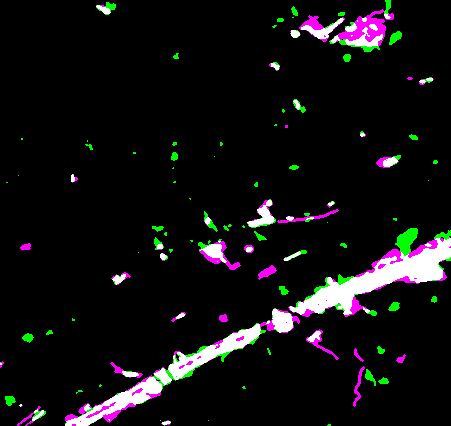,width=0.8\linewidth}}
    \centerline{(f) FC-Siam-diff.}\medskip
  \end{minipage}

  \caption{Illustrative results on the \textit{montpellier} test case of the OSCD dataset using all 13 color channels. In images (d), (e), and (f) white means true positive, black means true negative, green is false positive, and magenta is false negative.}
  \label{fig:sent_comp}
\end{figure*}

\begin{figure*}[ht]

  \begin{minipage}[b]{0.13\linewidth}
    \centering
    \centerline{\epsfig{figure=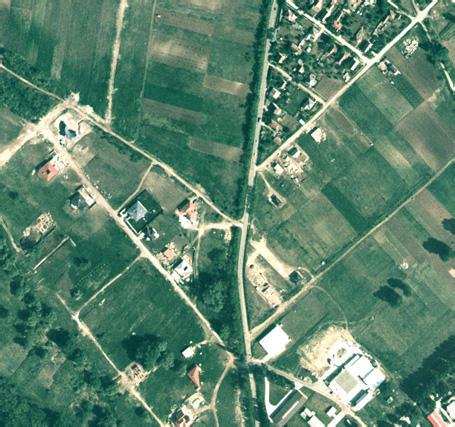,width=0.8\linewidth}}
    \centerline{(a) Image 1.}\medskip
  \end{minipage}
  \hfill
  \begin{minipage}[b]{0.13\linewidth}
    \centering
    \centerline{\epsfig{figure=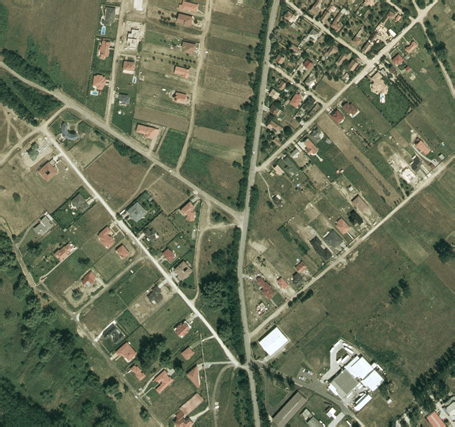,width=0.8\linewidth}}
    \centerline{(b) Image 2.}\medskip
  \end{minipage}
  \hfill
  \begin{minipage}[b]{0.13\linewidth}
    \centering
    \centerline{\epsfig{figure=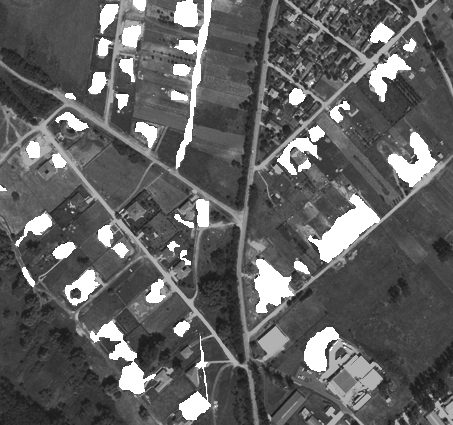,width=0.8\linewidth}}
    \centerline{(c) Ground truth.}\medskip
  \end{minipage}
  \hfill
  \begin{minipage}[b]{0.13\linewidth}
    \centering
    \centerline{\epsfig{figure=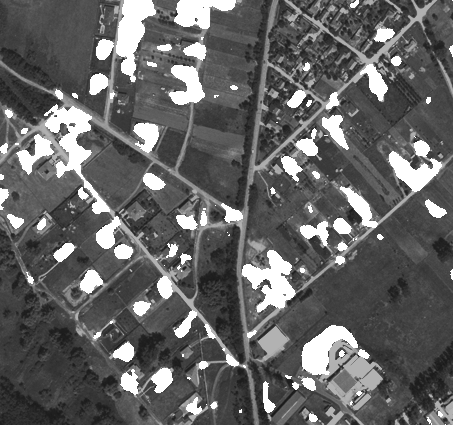,width=0.8\linewidth}}
    \centerline{(d) DSCN.}\medskip
  \end{minipage}
  \hfill
  \begin{minipage}[b]{0.13\linewidth}
    \centering
    \centerline{\epsfig{figure=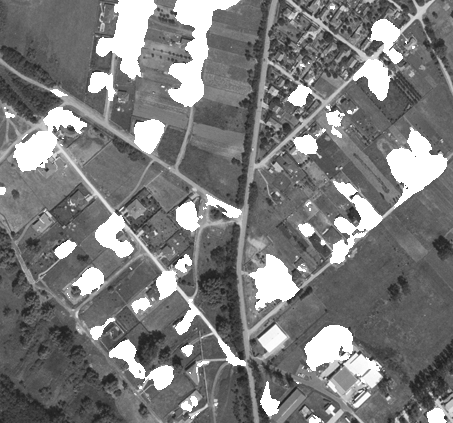,width=0.8\linewidth}}
    \centerline{(e) FC-EF.}\medskip
  \end{minipage}
  \hfill
  \begin{minipage}[b]{0.13\linewidth}
    \centering
    \centerline{\epsfig{figure=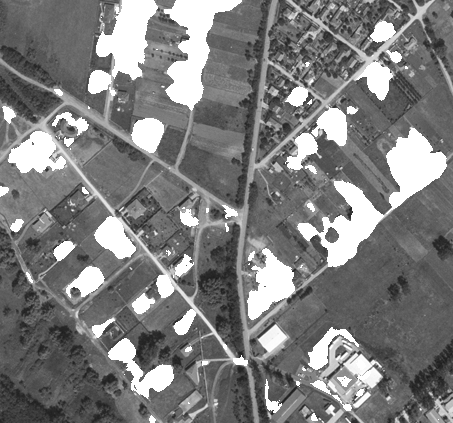,width=0.8\linewidth}}
    \centerline{(f) FC-Siam-conc.}\medskip
  \end{minipage}
  \hfill
  \begin{minipage}[b]{0.13\linewidth}
    \centering
    \centerline{\epsfig{figure=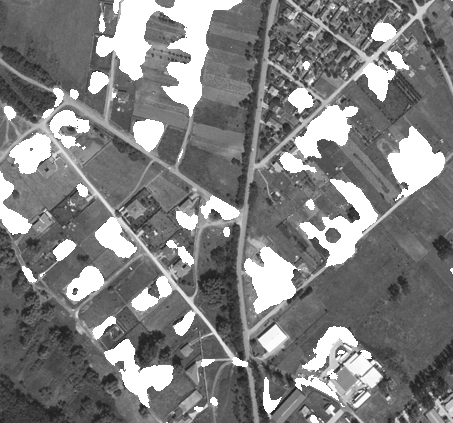,width=0.8\linewidth}}
    \centerline{(g) FC-Siam-diff.}\medskip
  \end{minipage}

  \begin{minipage}[b]{0.13\linewidth}
    \centering
    \centerline{\epsfig{figure=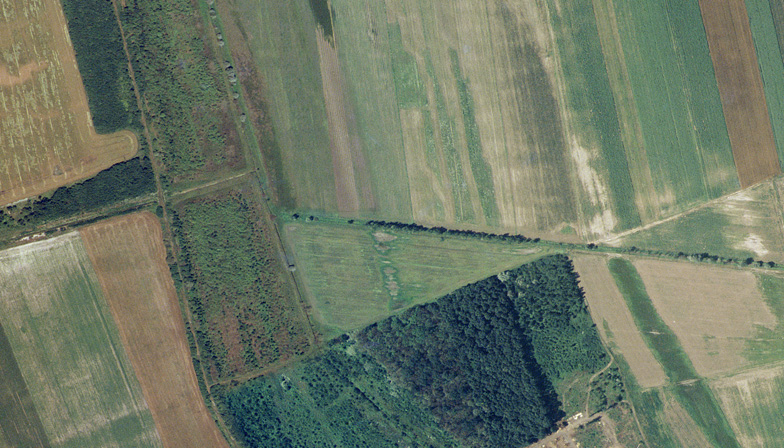,width=0.8\linewidth}}
    \centerline{(h) Image 1.}\medskip
  \end{minipage}
  \hfill
  \begin{minipage}[b]{0.13\linewidth}
    \centering
    \centerline{\epsfig{figure=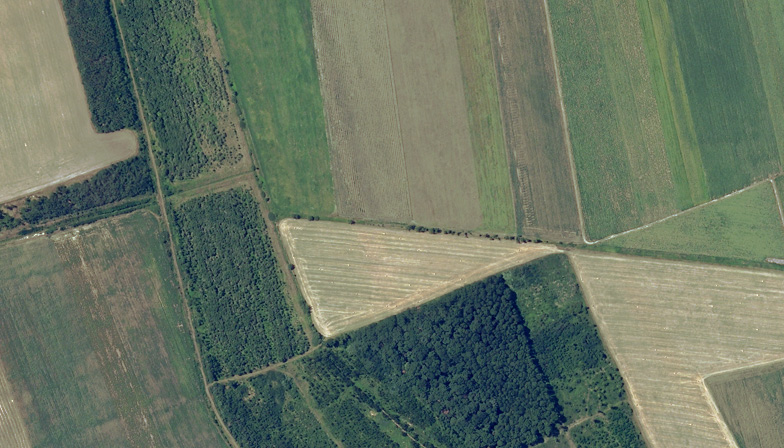,width=0.8\linewidth}}
    \centerline{(i) Image 2.}\medskip
  \end{minipage}
  \hfill
  \begin{minipage}[b]{0.13\linewidth}
    \centering
    \centerline{\epsfig{figure=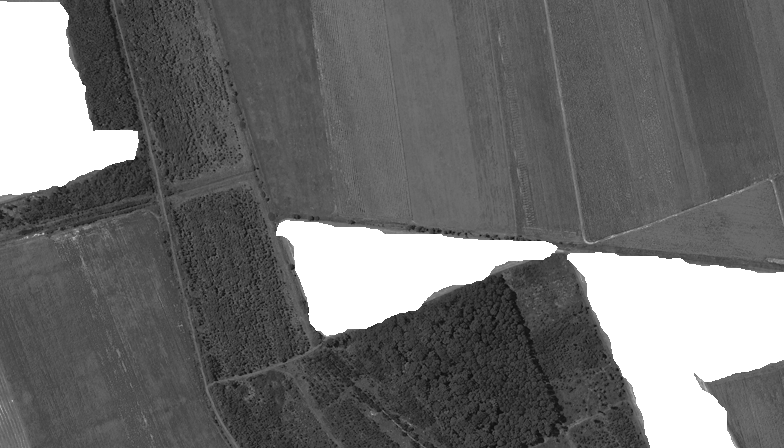,width=0.8\linewidth}}
    \centerline{(j) Ground truth.}\medskip
  \end{minipage}
  \hfill
  \begin{minipage}[b]{0.13\linewidth}
    \centering
    \centerline{\epsfig{figure=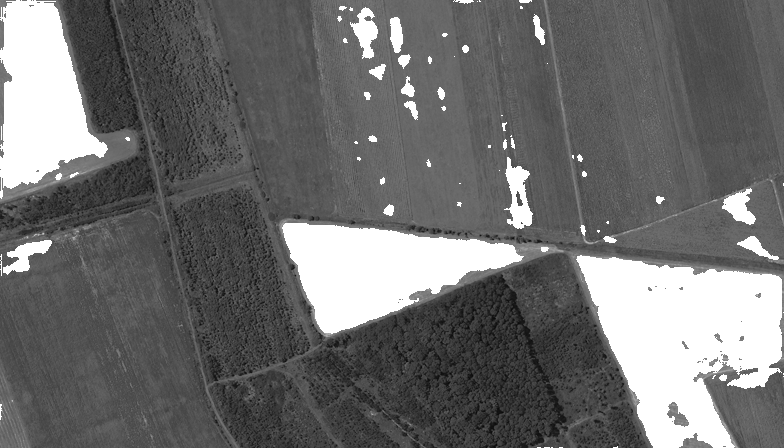,width=0.8\linewidth}}
    \centerline{(k) DSCN.}\medskip
  \end{minipage}
  \hfill
  \begin{minipage}[b]{0.13\linewidth}
    \centering
    \centerline{\epsfig{figure=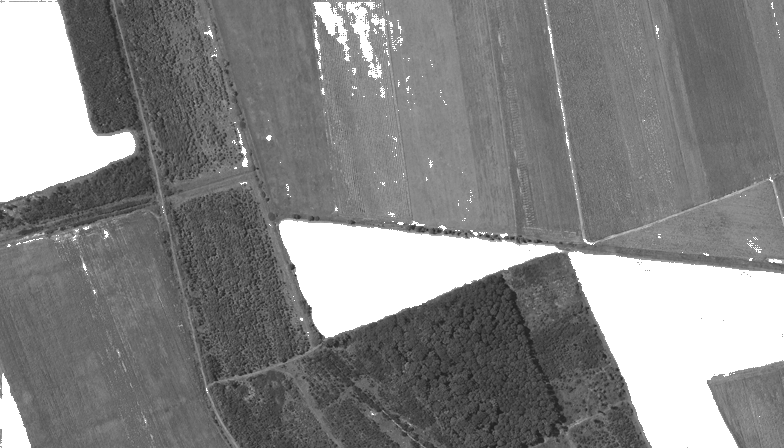,width=0.8\linewidth}}
    \centerline{(l) FC-EF.}\medskip
  \end{minipage}
  \hfill
  \begin{minipage}[b]{0.13\linewidth}
    \centering
    \centerline{\epsfig{figure=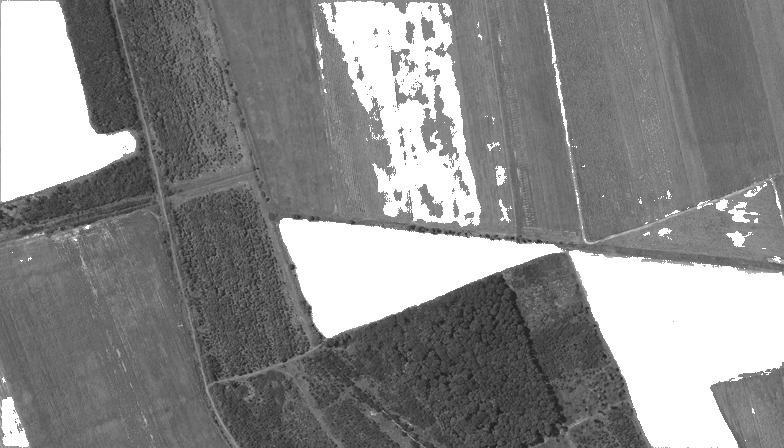,width=0.8\linewidth}}
    \centerline{(m) FC-Siam-conc.}\medskip
  \end{minipage}
  \hfill
  \begin{minipage}[b]{0.13\linewidth}
    \centering
    \centerline{\epsfig{figure=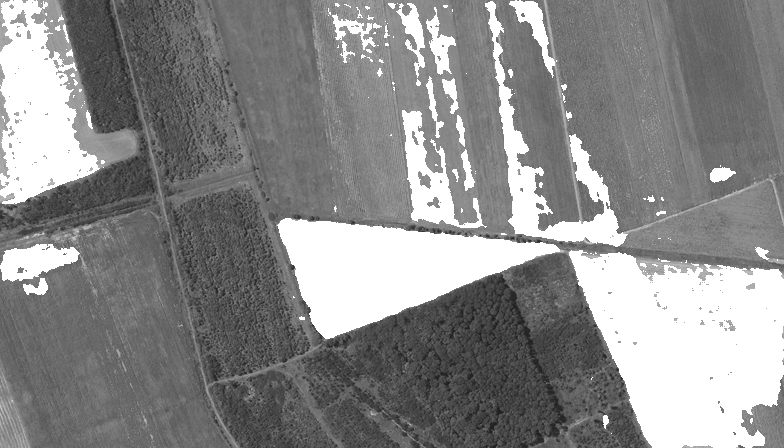,width=0.8\linewidth}}
    \centerline{(n) FC-Siam-diff.}\medskip
  \end{minipage}

  \caption{Comparison between the results obtained by the method presented in \cite{zhan2017change} and the ones described in this paper on the Air Change dataset. First row: Szada/1, second row: Tiszadob/3.}
  \label{fig:ac_comp}
\end{figure*}

\begin{table}
\centering
\begin{tabular}{cc|c|c|c|c}

Data & Network & Prec. & Recall & Global & F1\\
\hline \hline
 \parbox[t]{2mm}{\multirow{5}{*}{\rotatebox[origin=c]{90}{OSCD-3 ch.}}} & Siam.~\cite{daudt2018urban}        & 21.57 & 79.40  & 76.76 & 33.85 \\
         & EF~\cite{daudt2018urban} & 21.56 & \textbf{82.14} & 83.63 & 34.15 \\ \cline{2-6}
         & FC-EF        & 44.72 & 53.92 & 94.23 & \textbf{48.89} \\ 
         & FC-Siam-conc & 42.89 & 47.77 & 94.07 & 45.20 \\ 
         & FC-Siam-diff & \textbf{49.81} & 47.94 & \textbf{94.86} & 48.86 \\
\hline \hline 
\parbox[t]{2mm}{\multirow{5}{*}{\rotatebox[origin=c]{90}{OSCD-13 ch.}}}   & Siam.~\cite{daudt2018urban}        & 24.16 & \textbf{85.63} & 85.37 & 37.69 \\
         & EF~\cite{daudt2018urban}     & 28.35 & 84.69 & 88.15 & 42.48 \\ \cline{2-6}
         & FC-EF        & \textbf{64.42} & 50.97 & \textbf{96.05} & 56.91 \\ 
         & FC-Siam-conc & 42.39 & 65.15 & 93.68 & 51.36 \\ 
         & FC-Siam-diff & 57.84 & 57.99 & 95.68 & \textbf{57.92} \\  \hline\hline 
\parbox[t]{2mm}{\multirow{6}{*}{\rotatebox[origin=c]{90}{Szada/1}}}    
         & DSCN~\cite{zhan2017change} & 41.2 & 57.4 & NA & 47.9 \\ 
         & CXM~\cite{benedek2009change} & 36.5 & 58.4 & NA & 44.9 \\ 
         & SCCN~\cite{liu2016deep} & 24.4 & 34.7 & NA & 28.7 \\  \cline{2-6}
         & FC-EF        & \textbf{43.57} & 62.65 & \textbf{93.08} & 51.40 \\ 
         & FC-Siam-conc & 40.93 & 65.61 & 92.46 & 50.41 \\ 
         & FC-Siam-diff & 41.38 & \textbf{72.38} & 92.40 & \textbf{52.66} \\ \hline\hline 
\parbox[t]{2mm}{\multirow{6}{*}{\rotatebox[origin=c]{90}{Tiszadob/3}}} 
         & DSCN~\cite{zhan2017change} & 88.3 & 85.1 & NA & 86.7 \\ 
         & CXM~\cite{benedek2009change} & 61.7 & 93.4 & NA & 74.3 \\ 
         & SCCN~\cite{liu2016deep} & \textbf{92.7} & 79.8 & NA & 85.8 \\ \cline{2-6}
         & FC-EF        & 90.28 & 96.74 & \textbf{97.66} & \textbf{93.40} \\ 
         & FC-Siam-conc & 72.07 & \textbf{96.87} & 93.04 & 82.65 \\ 
         & FC-Siam-diff & 69.51 & 88.29 & 91.37 & 77.78 \\ \hline
\end{tabular}
\caption{Evaluation metrics on change detection datasets, in percent.}
\label{tab:results}
\end{table}

To evaluate the proposed methods we used two change detection datasets openly available. The first one is the Onera Satellite Change Detection dataset~\cite{daudt2018urban} (OSCD), and the second is the Air Change Dataset~\cite{benedek2009change} (AC). AC contains RGB aerial images, while OSCD contains multispectral satellite images. The networks were also tested using only the RGB layers of the OSCD dataset. The two classes (change and no change) were assigned weights inversely proportional to the number of examples in each one. The available data was augmented by using all possible flips and rotations multiple of 90 degrees to the training patches. Dropout was used to help avoid overfitting during training. 
All experiments were done using PyTorch and with an Nvidia GTX1070 GPU.


On the OSCD dataset, we split the data in train and test groups as recommended by the dataset's creators: fourteen images were used for training and ten images were used for testing. For the AC dataset, we followed the data split that was proposed in \cite{zhan2017change}: the top left 748x448 rectangle of the Szada-1 and Tiszadob-3 images were extracted for testing, and the rest of the data for each location (another 6 and 4 images, respectively) was used for training. This allowed direct comparison to three CD algorithms. Each location (Szada and Tiszadob) was treated completely separately as two different datasets, and the images named "Archieve" were ignored, since it comprises only one image pair which is not enough data to train the models presented in this paper.

Table~\ref{tab:results} contains the quantitative evaluation of the proposed CD architectures, along the same measures of other state-of-the-art methods. The Early Fusion and Siamese methods presented in \cite{daudt2018urban} were used for comparison for the OSCD dataset. For the AC dataset, the methods user for comparison were DSCN~\cite{zhan2017change}, CXM~\cite{benedek2009change}, and SCCN~\cite{liu2016deep}, using the values claimed by Zhan et al. in \cite{zhan2017change}. The table contains the precision, recall and F1 rate from the point of view of the "change" class, as well as the global precision when available.

The results relative to the OSCD dataset show that the methods proposed in this paper far outperform the ones proposed in \cite{daudt2018urban}, which is the precursor of this work. While the patch based methods achieve good recall rates, they do very poorly on the precision metric, which also reduces the F1 rate. Inference time of the fully convolutional architectures was under 0.1~s per image for all our test cases, while the patch based voting approach took several minutes to predict a change map for each image pair. On this dataset, the FC-EF and FC-Siam-diff architectures obtained the best F1 scores, while the results obtained by FC-Siam-conc were still clearly superior to the ones obtained by the patch based approaches. An illustration of our results on this dataset can be found in Fig.~\ref{fig:sent_comp}.

The results obtained on the AC dataset also show the superiority of our method compared to other previous methods. For the Szada/1 case, all proposed architectures outperformed the other methods used for comparison in the F1 metric, the FC-Siam-diff being once again the best architecture. For the Tiszadob/3 case, the best F1 score was obtained by our FC-EF architecture, while the other architectures were outperformed by DSCN and SCCN. Once again the inference time of the fully convolutional architectures were below 0.1~s, which is a speedup of over 500x compared to the processing time of 50~s claimed by Zhan et al.~\cite{zhan2017change} for the SCCN method on a very similar setup. The results for these cases can be viewed in Fig.~\ref{fig:ac_comp}.



In these tests, the FC-Siam-diff architecture seems to be the most suited for change detection, followed closely by FC-EF. This is, we believe, due to three main factors which make this network especially suited for this problem. First, fully convolutional networks were developed with the express purpose of dealing with dense prediction problems, such as CD. Second, the Siamese architecture imbues into the system an explicit comparison between two images. Finally, the difference skip connections also explicitly guides the network to compare the differences between the images, in other words, to detect the changes between the two images. The FC-EF architecture, which is more generic, also achieved excellent results despite having to learn the heuristics used for developing the FC-Siam-diff architecture from the training data.

The significant speedup of these FCNNs with no loss in performance compared to previous CD methods is a step towards efficient processing of the massive streams of Earth observation data which are available through programs like Copernicus and Landsat. These programs monitor very large areas with a high revisit rate. Deployment of such systems in conjunction with the methods proposed in this paper would enable accurate and fast worldwide monitoring.





\section{Conclusion}
\label{sec:conclusion}

In this paper we presented three fully convolutional networks trained end-to-end from scratch that surpassed the state-of-the-art in change detection, both in accuracy and in inference speed without the use of post-processing. Most notably, the fully convolutional encoder-decoder paradigm was modified into a Siamese architecture, using skip connections to improve the spatial accuracy of the results.

A natural extension of the work presented on this paper would be to evaluate how these networks perform at detecting semantic changes. It would also be interesting to test them with other image modalities (e.g. SAR), and to attempt to detect changes in image sequences. It is also likely that these networks would profit from training on larger datasets.




\newpage

\bibliographystyle{IEEEbib}
\bibliography{strings,refs}

\end{document}